\documentclass[letterpaper, 10 pt, conference]{ieeeconf}  

\IEEEoverridecommandlockouts                              
\usepackage{graphics} 
\usepackage{epsfig} 
\usepackage{amsmath} 
\usepackage{amssymb}  

\usepackage{url}
\usepackage[ruled, vlined, linesnumbered]{algorithm2e}
\usepackage{verbatim} 
\usepackage{soul, color}
\usepackage{lmodern}
\usepackage{fancyhdr}
\usepackage[utf8]{inputenc}
\usepackage{fourier} 
\usepackage{array}
\usepackage{makecell}
\usepackage{balance}

\usepackage[noend]{algpseudocode}

\SetNlSty{large}{}{:}

\makeatletter

\newcommand{\Rom}[1]{\expandafter\@slowromancap\romannumeral #1@}
\makeatother

\title{\LARGE \bf
Empirical Analysis of Anomaly Detection on Hyperspectral Imaging Using Dimension Reduction Methods
}

\author{Dongeon Kim$^{\ast}$\thanks{$^{\ast}$This work was done during an internship of the first author at SK Planet Co., Ltd.}, YeongHyeon Park$^{\dagger}$\thanks{$^{\dagger}$Corresponding author (yeonghyeon@sk.com)}
\\ SK Planet Co., Ltd. \\
}

\begin{document}

\maketitle
\thispagestyle{plain}
\pagestyle{plain}

\begin{abstract}
Recent studies try to use hyperspectral imaging (HSI) to detect foreign matters in products because it enables to visualize the invisible wavelengths including ultraviolet and infrared. Considering the enormous image channels of the HSI, several dimension reduction methods—e.g., PCA or UMAP—can be considered to reduce but those cannot ease the fundamental limitations, as follows: (1) latency of HSI capturing. (2) less explanation ability of the important channels. In this paper, to circumvent the aforementioned methods, one of the ways to channel reduction, on anomaly detection proposed HSI. Different from feature extraction methods (i.e., PCA or UMAP), feature selection can sort the feature by impact and show better explainability so we might redesign the task-optimized and cost-effective spectroscopic camera. Via the extensive experiment results with synthesized MVTec AD dataset, we confirm that the feature selection method shows 6.90$\times$ faster at the inference phase compared with feature extraction-based approaches while preserving anomaly detection performance. Ultimately, we conclude the advantage of feature selection which is effective yet fast.
\end{abstract}



\section{Introduction}
Anomaly detection has been one of the main research in machine learning, which is widely used in cyber-intrusion~\cite{ref1}, medical~\cite{ref2,ref3}, and industrial anomaly detection~\cite{ref4,ref5,ref6}. Recently, due to the development of equipment and to overcome the limitations of RGB imagery systems, many studies try to use hyperspectral imaging (HSI) that enables to represent abundant information by capturing spatial and spectral information even invisible to the human eye to detect foreign matters~\cite{ref7,ref8,ref9,ref10}. However, most of the studies are usually focused on detection performance but overlook model latency~\cite{ref11,ref12}. Specifically, when addressing HSI, they only consider enhancing detection performance although requires more time than RGB imagery systems because of capturing a wide range of spectra including ultraviolet (UV) and infrared (IR). Some of the studies have shown the effectiveness of dimension reduction methods (DRMs) named feature extraction (FE) with principal component analysis (PCA)~\cite{ref13,ref14} or uniform manifold approximation and projection (UMAP)~\cite{ref15}. 

Despite the excellent performance of FE, it suffers from the problems following: (1) It cannot reduce the equipment cost and collection time of HSI due to the whole spectrum in HSI without reducing the filter of the spectroscopic camera. (2) Moreover, it is difficult to explain which channel is the important than other. Refer that since the spectroscopic camera records the target spectrum in a line-scanning manner~\cite{ref16}, the total capturing time will be shortened when the number of filters is smaller.

In this paper, we revisit feature selection (FS) methods that one of the conventional DRMs that can directly reduce the equipment cost and explain important channels. Then, we compared to FS and FE methods. Referring to the DRMs that might induce degraded performance due to information loss, we desire to find the method that minimizes the discrepancy (i.e., trade-off) between performance and latency. Since it is difficult to disclose HSI data of a specific manufacturing process for security reasons, we use the MVTec AD dataset to configure and experiment with synthetic HSI data. As extensive experiments on the benchmark dataset, we confirmed well-chosen important spectrum methods and demonstrate equivalent and preserve detection performance as well as more than 6.90× as fast at the inference phase. Ultimately, we contribute to reducing the latency and equipment cost for the given process via important spectrums selection within numerous wavelengths for tailored designing the spectroscopic camera. Moreover, we provide insight to select the methods suitable for the purpose of the researcher or industrial manager.

\begin{figure*}
    \begin{center}
        \includegraphics[width=0.9\linewidth]{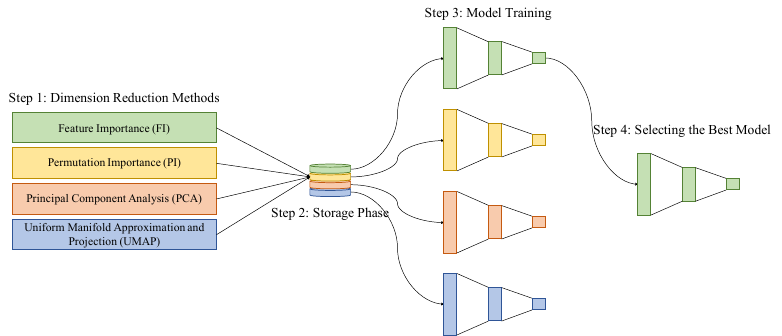}
    \end{center}
    \caption{An illustration of a schematic framework for comparison of anomaly detection performance with dimension reduction methods.}
    \label{fig:fig1}
\end{figure*}

\section{Method}
To compare performance among DRMs, we leverage a strategy that consists of pre-processing phase and anomaly detection phase. Our approach consists of four steps. Schematic our framework is shown in Figure~\ref{fig:fig4} and more detail for our framework algorithms is shown in Algorithm~\ref{algo:algo1}.

\begin{algorithm}
\caption{\scriptsize{The DRM-based Anomaly Detection Algorithm.}}
    \begin{algorithmic}[1]
        \State Inputs: $x^{a} \in \mathbb{R}^{C\times{}H\times{}W}$, $N^{b}$, $D^{c}$

        \SetKwInOut{TPhase}{Training Phase}
        \hspace{-36pt} \textbf{\TPhase{}}
        
        \State Select FS or FE 
        \State \textbf{if} FE \textbf{then} 
            \State $\;$ $C_{sorted} = FS(x)$, sorted by the importance
            \State $\;$ \textbf{return} $x_{reduce} = x(C_{sorted}[:N])$
        \State \textbf{end if} 

        \State \textbf{if} FE(N) \textbf{then} 
            \State $\;$ $FE.fit(x)$
            \State $\;$ \textbf{return} $x_{reduce} = FE.transform(x)$
        \State \textbf{end if}
        \State \textbf{for all} $x_{reduce}$ \textbf{do}
            \State $\;$ $\hat{y_{i}}=D(x_{reduce_{i}})$
            \State $\;$ $L_{CE}=CrossEntropy(x_{reduce_{i}}, \hat{x_{i}})$
            \State $\;$ Update weight based on optimizer
            \State $\;$ \textbf{until} coverage $L_{CE}$
        \State \textbf{end for}
        \State \textbf{save} $D$

        \SetKwInOut{IPhase}{Inference Phase}
        \hspace{-36pt} \textbf{\IPhase{}}

        \State Select FS or FE
        \State \textbf{if} FS \textbf{then} 
            \State $\;$ load $C_{sorted}$
            \State $\;$ \textbf{return} $x_{reduce} = x(C_{sorted}[:N])$
        \State \textbf{end if}
        \State \textbf{if} FE \textbf{then} 
            \State $\;$ load fitted $FE$
            \State $\;$ \textbf{return} $x_{reduce} = FE.transform(x)$
        \State \textbf{end if}
        \State load pre-trained $D$
        \State $\hat{y}=D(x_{reduce})$
        \State Compute \textbf{Anomaly Score$(y, \hat{y})$} and \textbf{AUROC$(y, \hat{y})$}
    \end{algorithmic}
    \label{algo:algo1}
\end{algorithm}

\subsection{Preprocessing}
We define several dimension reduction methods to compare the effectiveness. For the FS, we exploit the feature importance (FI) and permutation importance (PI). In addition, PCA is used for a FE approach. The UMAP is not used in this study because its low throughput. All equations corresponding to the FI, PI, and PCA are below.

\begin{equation}
    \begin{aligned}
        FI = \sum_{i=1}^{C} f_{i} (1-f_{i})
    \end{aligned}
    \label{eq:eq1}
\end{equation}

\begin{equation}
    \begin{aligned}
        PI = s - \frac{1}{K} \sum_{k=1}^{K} s_{k,j}
    \end{aligned}
    \label{eq:eq2}
\end{equation}

\begin{equation}
    \begin{aligned}
        PCA = Q\Lambda{}Q^{T}
    \end{aligned}
    \label{eq:eq3}
\end{equation}

\subsection{Anomaly detection}
The results of pre-processing phase will be fed into anomaly detection model $f:X \rightarrow Y \in\{0,1\}^n$. Referring to previous anomaly detection studies, we adopt a basic end-to-end supervised anomaly scoring model for anomaly detection because it is simple yet effective~\cite{ref17,ref18}. Although we adopted Auto-Encoder techniques, also can use anomaly detection methods such as VAE~\cite{ref19}, U-Net~\cite{ref20}, PaDiM~\cite{ref21}, or etc.

The anomaly scoring model is trained with cross-entropy loss as a binary classification manner as below.

\begin{equation}
    \begin{aligned}
        \mathcal{L}_{CE}(Y, \hat{Y}) = -\frac{1}{N} \sum_{i=1}^{N}y_{i}log(q_{i}) + (1-y_{i})log(1-q_{i})
    \end{aligned}
    \label{eq:eq4}
\end{equation}

According to the above objective function, the neural network will be trained to indicate that the closer to 1 is, the more anomalous it is. Thus, the output $\hat{Y}$ can be directly used as an anomaly score in the inference phase. For quantitative evaluation, we adopt area under the receiver operating characteristics (AUROC) as the evaluation metric~\cite{ref22,ref23}.

\begin{figure}
    \begin{center}
        \includegraphics[width=0.95\linewidth]{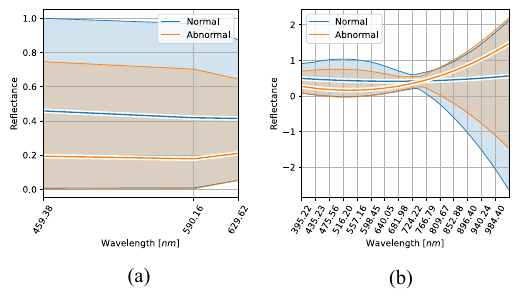}
    \end{center}
    \caption{Each (a) and (b) show the sample spectrum of original RGB image and synthesized HSI, respectively. Original spectrum only can represent three wavelengths. The spectrum of HSI that synthesized by interpolation is consisted with 300 wavelengths.}
    \label{fig:fig2}
\end{figure}

\begin{figure}
    \begin{center}
        \includegraphics[width=0.95\linewidth]{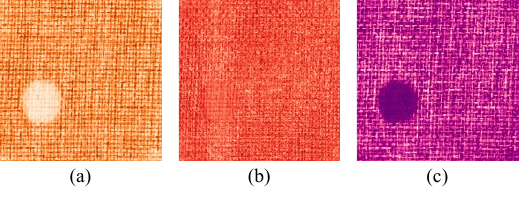}
    \end{center}
    \caption{Single channel images extracted from synthesized HSI. (a), (b), (c) indicate the orange (660.98nm), red (751.85nm), and near-infrared ranges (1015.52nm), respectively. Since the capability to identify the anomalous region is different for each channel, we should selectively use the important channels.}
    \label{fig:fig3}
    \vspace{-0.1cm}
\end{figure}

\begin{table}[ht]
    \centering
    \caption{Experimental settings}
    \begin{tabular}{ll}
        \hline
            \textbf{Hyperparameter} & \textbf{Value} \\       
        \hline
        \hline
            Optimizer & Adam \\
            Learning rate & 1e-5 \\ 
            Batch size & 8 \\ 
            Target \# channels (w/ Top@N) & 6 \\
            Scalar & Min-max scalar \\
        \hline
    \end{tabular}
    \label{tab:table1}
\end{table}

\begin{table}[ht]
    \centering
    \caption{Comparison of Model Performance}
    \begin{tabular}{lrrrr}
        \hline
            \textbf{Class} & \textbf{Origin} & \textbf{FI} & \textbf{PI} & \textbf{PCA} \\       
        \hline
        \hline
            Carpet & 73.3 & 86.7 & 86.7 & 60.0 \\
            Leather & 100.0 & 100.0 & 100.0 & 100.0 \\
            Tile & 76.7 & 83.3 & 83.3 & 73.3 \\
            Wood & 100.0 & 83.3 & 83.3 & 83.3 \\
        \hline
            Avg. & 87.5 & 88.3 & 88.3 & 79.2 \\
        \hline
    \end{tabular}
    \label{tab:table2}
\end{table}

\begin{table}[ht]
    \centering
    \caption{Time consumption of inference with and without DRMs}
    \begin{tabular}{lrrr}
        \hline
            \textbf{Method} & \textbf{Origin (w/o DRM)} & \textbf{FI/PI} & \textbf{PCA} \\       
        \hline
        \hline
            Time (sec/sample) & 0.855 & 0.124 & 0.687 \\
        \hline
    \end{tabular}
    \label{tab:table3}
\end{table}

\section{Results}
\subsection{Dataset and evaluation metrics}
In this paper, we use four texture categories in MVTec AD dataset~\cite{ref24}, which is widely used benchmark dataset in anomaly detection tasks. However, it has formed RGB format. To treat HSI, we build a synthetic dataset based on MVTec AD dataset. Specifically, we interpolate three RGB channels into 300 points hyperspectral channels corresponding from UV to IR. A sample results of HSI synthesis is shown in Figure~\ref{fig:fig2} and its corresponding images are shown in Figure~\ref{fig:fig3}. 

Referring to the single-channel images shown in Figure~\ref{fig:fig3}, extracted from three wavelength points, the anomalous region can be easily recognized at 660.98nm and 1015.52nm, but cannot at 751.85nm. Thus, to improve the system efficiency, we should selectively use the important channels that can significantly affect to detect anomalies among all 300 channels.

\subsection{Experiment settings}
We resize all the images to 256×256 that included selected subtask of MVTec AD then synthesize the HSI. Sequentially, we feed the synthetic HSI into convolutional neural network-based anomaly scoring model. We implemented in PyTorch~\cite{ref25} and scikit-learn~\cite{ref26}. More detail about experiment settings is shown in Table~\ref{tab:table1}.

\begin{figure}
    \begin{center}
        \includegraphics[width=0.95\linewidth]{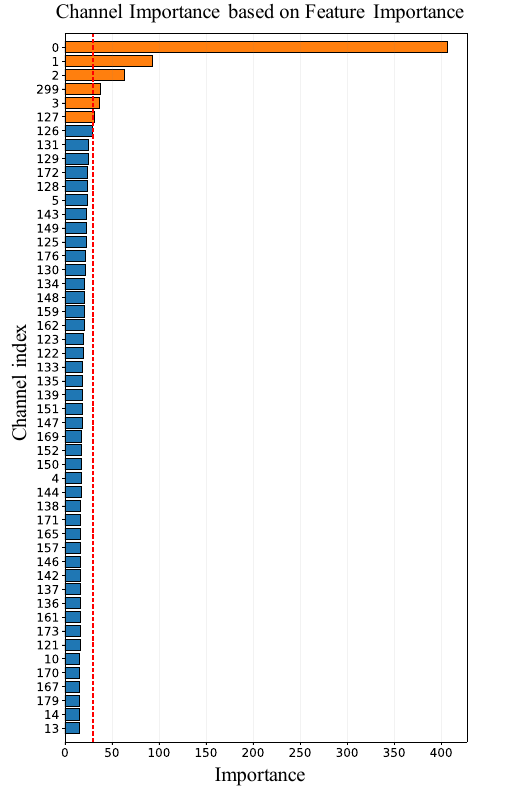}
    \end{center}
    \caption{A result of measured feature importance for carpet dataset. Within all channels, we discard all channels except top six important channels for reducing channel dimension.}
    \label{fig:fig4}
\end{figure}

\subsection{Comparison of model performance}
As experiment results for comparison of detection performance with DRMs, we confirm that dimension reduction methods are shown equivalent performance in leather dataset and even outperform using the original HSI (i.e., using 300 channels) in carpet and tile dataset. However, in wood dataset, which is shown degrade detection performance. Specifically, in DRMs, FS methods are shown equivalent performance in leather and wood datasets, and outperform FE methods in carpet and tile datasets, more detail is shown in Table~\ref{tab:table2}. Refer that an example FI result is shown in Figure~\ref{fig:fig4}. 

\subsection{Latency analysis}
As experiment results for latency analysis, we measure inference time with and without DRMs. We confirm that throughput is 6.90× faster when FS, including FI and PI, is applied than the original HSI case at the inference phase as shown in Table~\ref{tab:table3}.

\section{Conclusion}
In this paper, we revisit FS methods to improve the efficiency of the HSI-based anomaly detection systems. The HSI, which we use, is structured with 300 channels that should slow down not only target image capturing but also training and inference of anomaly detection model. 

Referring that the importance within all channels to determine abnormality is different from each other, all the FS methods will help to reduce computational burden by discarding useless channels. In fact, our comparative experiments show that the FS methods improve the throughput of the anomaly detection model while preserving the detection performance. Furthermore, some of the FS methods show superior detection performance than when using full channels and using FE methods. Thus, we expect that the FS methods fundamentally enhance the cost-effectiveness to construct an anomaly detection model and help to design the capturing device using important channels. As the result, we can explain what is important channels in HSI wavelengths by using FS methods.


\balance

\end{document}